\documentclass[letterpaper, 10 pt, journal, twoside]{ieeeconf} 
\IEEEoverridecommandlockouts

\IEEEoverridecommandlockouts          

\usepackage{cite}
\usepackage{url}
\usepackage{amsmath,amssymb,amsfonts}
\usepackage{graphicx}
\usepackage{booktabs}
\usepackage{multirow}
\usepackage{subcaption}
\usepackage{algorithm}
\usepackage{algpseudocode}
\usepackage{textcomp}
\usepackage{xcolor}
\usepackage{longtable}  
\usepackage{stfloats}   
\usepackage{makecell}
\usepackage{placeins}   
\usepackage[utf8]{inputenc}
\usepackage{mathtools}
\usepackage{bm}
\usepackage{caption}
\usepackage{subcaption}

\usepackage{tabularx}
\newcolumntype{Y}{>{\raggedright\arraybackslash}X}
\title{\LARGE \bf
InDRiVE: Reward-Free World-Model Pretraining for Autonomous Driving via Latent Disagreement 
}

\author{Feeza Khan Khanzada and Jaerock Kwon
\thanks{*This work was supported in part by the National Science
Foundation (NSF) under Grant 2500638.}
\thanks{Feeza Khan Khanzada and Jaerock Kwon are with the Department of Electrical and Computer Engineering, University of Michigan-Dearborn, 4901 Evergreen Rd, Dearborn, MI 48128, United States. {\tt\small \{feezakk, jrkwon\}@umich.edu}}%
}

\begin{document}

\thispagestyle{empty}
\pagestyle{empty}

\maketitle
\thispagestyle{empty}
\pagestyle{empty}

\begin{abstract}

Model-based reinforcement learning (MBRL) can reduce interaction cost for autonomous driving by learning a predictive world model, but it typically still depends on task-specific rewards that are difficult to design and often brittle under distribution shift. This paper presents InDRiVE, a DreamerV3-style MBRL agent that performs reward-free pretraining in CARLA using only intrinsic motivation derived from latent ensemble disagreement. Disagreement acts as a proxy for epistemic uncertainty and drives the agent toward under-explored driving situations, while an imagination-based actor–critic learns a planner-free exploration policy directly from the learned world model. After intrinsic pretraining, we evaluate zero-shot transfer by freezing all parameters and deploying the pretrained exploration policy in unseen towns and routes. We then study few-shot adaptation by training a task policy with limited extrinsic feedback for downstream objectives (lane following and collision avoidance). Experiments in CARLA across towns, routes, and traffic densities show that disagreement-based pretraining yields stronger zero-shot robustness and robust few-shot collision avoidance under town shift and matched interaction budgets, supporting the use of intrinsic disagreement as a practical reward-free pretraining signal for reusable driving world models.


\end{abstract}

\section{Introduction}

Autonomous vehicles must handle many different situations, from new road layouts to changing traffic levels and weather. Reinforcement Learning (RL) is appealing in this setting because it can learn driving behaviors through trial and error. In practice, however, RL often depends on carefully designed reward functions and large amounts of task-specific experience. Designing driving rewards is time-consuming, and small reward design choices can lead to behaviors that work in one town or traffic setting but fail to transfer to another.

Model Based Reinforcement Learning (MBRL) can reduce the amount of experience needed by first learning a predictive model of the environment—effectively a learned simulator and then using that model to train a policy \cite{delavari_comprehensive_2025}. But many model-based driving agents still rely on task rewards throughout training. As a result, they tend to learn behaviors and representations that are tightly tied to a particular task, rather than knowledge that can be reused across tasks and environments \cite{khanzada_driving_2025}.

Curiosity-driven learning offers a different training signal. Instead of rewarding the agent for completing a specific task, the agent can be rewarded for exploring situations that are new or hard to predict. This leads to a central question for urban driving: \emph{can reward-free exploration alone produce a reusable driving model and a useful driving behavior, and can these transfer to downstream tasks with only a small amount of additional training?}

We study this question in the CARLA simulator \cite{Dosovitskiy17}, an open-source, high fidelity simulator designed for research in autonomous driving, providing realistic urban environments, sensor models, and configurable traffic to enable safe and reproducible experimentation. We propose \textbf{InDRiVE} (Intrinsic Disagreement-based Reinforcement for Vehicle Exploration), a model-based agent that learns by exploring. During pretraining, InDRiVE uses only an intrinsic reward based on \emph{disagreement} among multiple predictive models: when the models disagree about what will happen next, the agent treats that situation as interesting and explores it more. This encourages broad coverage of driving scenarios without using any task reward.

After reward-free pretraining, we evaluate transfer in two stages. First, we perform true zero-shot deployment: we freeze all parameters and run the exploration policy directly on unseen towns, routes, and traffic settings, without any further learning. Second, we perform few-shot specialization: we train a task policy for lane following and collision avoidance using only a small interaction budget (10k steps), while keeping the learned dynamics model fixed. This setup tests whether the pretrained model is broadly useful and whether it speeds up learning on safety-relevant tasks.

Finally, to isolate the effect of the intrinsic objective, we compare disagreement-based curiosity against two widely used intrinsic signals, Intrinsic Curiosity Module (ICM) \cite{pathak_curiosity-driven_2017} and Random Network Distillation (RND) \cite{burda_exploration_2018}, using the same backbone, replay setup, and interaction budgets.

This work makes the following contributions:


\begin{itemize}
    \item Reward-free driving pretraining in CARLA using a DreamerV3 world-model backbone and latent ensemble disagreement as the sole learning signal, producing a pretrained world model and an exploration policy without task rewards.
    \item A strict transfer protocol for driving world models: (i) true zero-shot deployment with all parameters frozen on a held-out town and standardized routes; (ii) few-shot task learning with a frozen dynamics model and a strict 10k-step extrinsic budget.
    \item A controlled intrinsic-objective study (disagreement vs. ICM vs. RND) under identical backbones and matched interaction budgets, quantifying cross-town generalization gaps and downstream safety behavior (collision avoidance).    
\end{itemize}

InDRiVE builds on disagreement-driven exploration in latent world models \cite{sekar_planning_2020} and adopts DreamerV3 as the backbone. Our contribution is the driving-specific reward-free pretraining + transfer evaluation protocol that isolates what intrinsic pretraining changes in zero-shot robustness and few-shot safety behavior under matched interaction budgets.

\section{Related Work}

MBRL improves data efficiency by learning a predictive model of the environment and using it to evaluate future outcomes before acting \cite{delavari_comprehensive_2025}. World-model approaches based on recurrent latent dynamics have
enabled imagination-based control in high-dimensional domains \cite{ha_recurrent_2018, hafner_dream_2020, hafner_mastering_2022}.
In autonomous driving, learned world models have been adopted to anticipate future scenes and traffic interactions in
simulators such as CARLA, including predictive individual world models \cite{gao_dream_2024} and
model-based imitation frameworks \cite{hu_model-based_2022}. Recent CARLA studies further emphasize the value of
high fidelity predictive modeling for downstream driving behaviors \cite{garg_imagine-2-drive_2025}. Despite these
advances, most driving agents remain coupled to task rewards or demonstrations, which can be costly to design and may
limit robustness under distribution shift \cite{delavari_comprehensive_2025}. This motivates reward-free pretraining strategies that learn
reusable driving representations and dynamics before specializing to specific tasks.

Intrinsic Motivation (IM) and curiosity-driven exploration have emerged as essential mechanisms for guiding agents in sparse-reward or high-dimensional environments, where extrinsic feedback is rare or too costly to define \cite{pathak_curiosity-driven_2017, burda_large-scale_2018-1}. IM provides agents with self-generated reward signals that encourage exploration, often by rewarding novelty, uncertainty, or prediction error \cite{meyer_possibility_1991, stadie_incentivizing_2015}. In model-based settings, epistemic uncertainty is often approximated using ensembles, where predictive disagreement provides an intrinsic reward that encourages visiting under-modeled regions of the state space \cite{sekar_planning_2020}. Notable curiosity-based approaches include the ICM and RND, both of which incentivize agents to visit unfamiliar or surprising states. Such methods have been successfully applied to robotic systems and video game domains, enabling agents to learn skills in the absence of dense external rewards \cite{burda_large-scale_2018-1, oudeyer_intrinsic_2007}. However, purely intrinsic exploration can lead agents to focus on irrelevant or noisy events, sparking interest in techniques that combine curiosity with additional constraints or memory mechanisms to ensure meaningful, goal-relevant exploration \cite{raileanu_ride_2020-1}.

In the realm of autonomous driving, previous research has primarily benefited from intrinsic motivation as a complementary signal rather than a primary training objective \cite{garg_imagine-2-drive_2025, gao_dream_2024, hu_model-based_2022}. Typical RL frameworks for driving rely on task-specific reward functions (e.g., measuring route progress, penalizing collisions, or encouraging lane-keeping) \cite{toromanoff_end--end_2020}, often augmented with a small curiosity bonus to expedite convergence. While this hybrid approach can alleviate some exploration hurdles, it still anchors the learned policy to a particular extrinsic objective, reducing its flexibility to generalize across tasks or conditions. Additionally, exploration in autonomous driving requires careful consideration of real-world feasibility; purely random or naive exploration is not viable in practice, further complicating the application of intrinsic rewards \cite{codevilla_exploring_2019}. As a result, exploration mechanisms in driving must be both informative and risk-aware, since novelty seeking alone can increase exposure to unsafe events without improving task-relevant competence \cite{codevilla_exploring_2019}.

Overall, most driving RL systems remain anchored to task-specific rewards or demonstrations, and intrinsic motivation is
typically used as an auxiliary exploration bonus rather than the sole learning signal \cite{gao_dream_2024, hu_model-based_2022, garg_imagine-2-drive_2025}.
Evidence for purely intrinsic, reward-free pretraining that yields a reusable driving world model and measurable zero-shot
transfer across unseen towns and traffic conditions remains limited. In particular, prior driving world-model studies typically rely on task rewards and do not report a protocol that (a) freezes
all parameters for zero-shot evaluation under town/layout shift and (b) compares intrinsic objectives under identical
world-model backbones and matched interaction budgets. We focus on this missing evaluation axis, using standardized
routes and traffic densities to quantify transfer and failure outcomes. We investigate whether latent ensemble disagreement can serve as a standalone pretraining driver and how efficiently downstream driving objectives can be recovered with a small fine-tuning budget.

\section{Methodology}
\label{sec:method}

\begin{figure*}[t]
  \centering
  \includegraphics[scale=0.55]{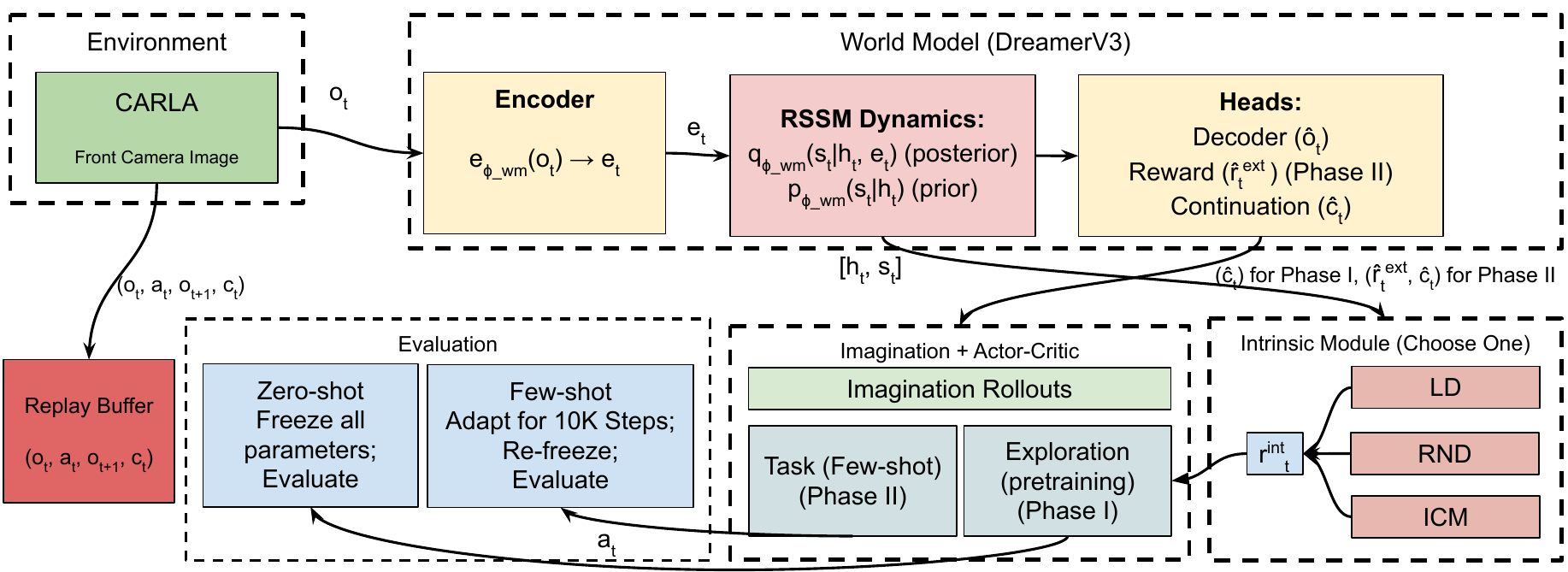}
  \captionsetup{font=footnotesize} 
  \caption{Overview of InDRiVE. During reward-free pretraining, the agent collects data using an exploration policy optimized only with intrinsic reward (LD/ICM/RND). The world model is trained from replay using reconstruction, dynamics, and continuation losses; no task reward is computed or used. Zero-shot transfer freezes all parameters and evaluates the pretrained exploration policy in unseen towns/routes. Few-shot fine-tuning then introduces task-specific rewards and trains a task policy for a small interaction budget; the dynamics model is kept frozen unless stated otherwise.
  }
  \vspace{-15pt}
  \label{fig:overall_methodology}
\end{figure*}

We propose \textbf{InDRiVE}, a DreamerV3-based \cite{hafner_mastering_2024} model-based driving agent for CARLA that uses ensemble-disagreement exploration~\cite{sekar_planning_2020}. Rather than introducing a new intrinsic objective, we focus on a controlled reward-free pretraining and transfer evaluation protocol in a critical driving domain, measuring (i) zero-shot robustness in held-out towns/routes and (ii) few-shot adaptation under smaller interaction budgets. As parity baselines, we implement ICM and RND on the same DreamerV3 backbone and replay buffer
and evaluate them under the same protocol. Figure~\ref{fig:overall_methodology} summarizes how
InDRiVE, ICM, and RND integrate into the DreamerV3 framework for intrinsic exploration.

\subsection{DreamerV3 world model}
At each step, the agent observes $o_t \in \mathcal{O}$ (front semantic-segmentation image) and executes $a_t \in \mathcal{A}$ (discrete control). We use a recurrent state-space model (RSSM) with deterministic hidden state $h_t$ and stochastic latent state $s_t$, and define the feature used by policies and intrinsic modules as $u_t=[h_t; s_t]$. We denote world-model parameters (encoder, RSSM dynamics, decoder, continuation) by $\phi_{wm}$ and the extrinsic reward-head parameters by $\phi_r$.


Let $e_t=e_\phi(o_t)$ be the encoder output. The deterministic RSSM transition $f_{\phi_{wm}}$ updates
\begin{align}
h_t &= f_{\phi_{wm}}(h_{t-1}, s_{t-1}, a_{t-1}).
\end{align}

During representation learning, we sample $s_t$ from the posterior
$q_{\phi_{\mathrm{wm}}}(s_t \mid h_t, e_t)$; during imagination rollouts,
we sample from the prior $p_{\phi_{\mathrm{wm}}}(s_t \mid h_t)$.



The model decodes observations $\hat{o}_t$, we write $\hat{o}_t$ for its reconstruction and predicts continuation $\hat{c_t}$ from $(h_t,s_t)$. We define $c_t \in \{0,1\}$ as the continuation label (1 if the episode continues at t+1, 0 if terminal), derived from the termination events in Sec \ref{sec:termination}.

During reward-free pretraining, intrinsic rewards $r_t^{\text{int}}$ are computed directly by the intrinsic module from latent features (LD/ICM/RND) and do not require a world-model reward head. During few-shot fine-tuning, we introduce an extrinsic reward head $p_{\phi_r}(r_t^{\text{ext}}\mid h_t,s_t)$ to support imagined-rollout training of the task policy. No task reward is computed, stored, or used during pretraining.


\begin{align}
\mathcal{L}_\text{model}^\text{pre}(\phi_{wm}) 
  &= \mathbb{E}_{q_{\phi_{wm}}}\Bigl[
        -\ln p_{\phi_{wm}}(o_t \mid h_t, s_t)
     \Bigr] \notag\\
  &\quad
     + \beta_{\text{KL}} \,\mathbb{E}_{q_{\phi_{wm}}}\Bigl[
        D_{\mathrm{KL}}\bigl(
            q_{\phi_{wm}}(s_t \mid h_t, e_t)
        \bigr. \notag\\
  &\qquad\qquad\qquad\bigl.\,\|\, 
            p_{\phi_{wm}}(s_t \mid h_t)
        \bigr)
     \Bigr] \notag\\
  &\quad
     + \lambda_{c}\,\mathbb{E}_{q_{\phi_{wm}}}\Bigl[
        -\ln p_{\phi_{wm}}(c_t \mid s_t, h_t)
     \Bigr],\label{eq:wm_pre} \\[0.25em]
\mathcal{L}_\text{model}^\text{fine}(\phi_r; \phi_{wm})
  &= \lambda_{r}\,\mathbb{E}_{q_{\phi_{wm}}}\Bigl[
        -\ln p_{\phi_r}(r^{\text{ext}}_t \mid s_t, h_t)
     \Bigr].\label{eq2}
\end{align}

In our few-shot setting, the pretrained world-model parameters $\phi_{wm}$ in $\mathcal{L}_\text{model}^\text{pre}$ (Eq.~\ref{eq:wm_pre}) are
kept frozen. During Phase II few-shot fine-tuning we optimize only the reward-head parameters $\phi_r$ in
$p_{\phi_r}(r_t^{ext}\mid h_t,s_t)$, with gradients stopped through $(h_t,s_t)$.


\subsection{Intrinsic Exploration in Latent Space}
\label{sec:intrinsic}
We drive exploration with \emph{epistemic} novelty measured as \emph{disagreement} in the world-model latent, and we implemented \emph{ICM} and \emph{RND} as parity baselines in the \emph{same} backbone and replay.

\subsubsection{Ensemble Disagreement for Intrinsic Exploration}
We use ensemble disagreement as an intrinsic reward.
We train $K$ lightweight predictors $\{w_k\}_{k=1}^K$ to predict the next stochastic latent from replay. Each predictor $w_k$ outputs a categorical distribution over the next discrete latent. We train $w_k$ on replay using cross-entropy to match the next-step posterior from the world model, stopping gradients into the world model.
Each predictor takes $(u_t,a_t)$ as input, where $u_t=[h_t;s_t]$ and $a_t$ is one-hot encoded action, and outputs
$s_{t+1,k}\in\mathbb{R}^d$ representing the predicted next-latent categorical probabilities.

DreamerV3 uses a discrete stochastic latent with $32$ categorical variables and $32$ classes each, so
$d=32\times 32=1024$. We compute intrinsic reward as the mean per-dimension ensemble variance:
\begin{equation}
r_t^{\text{int}} = \frac{1}{d} \sum_{i=1}^d \operatorname{Var}_k \left[ s_{t+1,k}^{(i)} (u_t, a_t) \right].
\label{eq:int_reward}
\end{equation}

To keep disagreement reflective of epistemic uncertainty rather than representation shaping, we stop gradients
into the world model when training $\{w_k\}$, and update the ensemble at the same frequency as world-model updates
using the same replay batches.

High disagreement indicates unexplored or uncertain regions, incentivizing the policy to gather data where the world model is less confident. As training progresses, this promotes coverage of diverse states and reduces model uncertainty in critical scenarios like driving.

\subsubsection{ICM baseline} \label{sec:intrinsic_icm}
ICM~\cite{pathak_curiosity-driven_2017} operates in feature space \(u_t = [h_t;s_t]\).
An inverse model predicts \(a_t\) from \((u_t,u_{t+1})\) and a forward model predicts \(u_{t+1}\) from \((u_t,a_t)\):
\begin{align}
\mathcal{L}^{\text{ICM}}_{\text{inv}} &= \mathrm{CE}\!\left(\hat{a}_t(u_t,u_{t+1}),\, a_t\right), \\
\mathcal{L}^{\text{ICM}}_{\text{fwd}} &= \big\| \hat{u}_{t+1}(u_t, a_t) - u_{t+1} \big\|_2^2.
\end{align}
The intrinsic reward is the normalized forward-prediction error:
\begin{align}
r^{\text{int}}_{\text{ICM},t} 
= \eta \cdot \mathrm{Norm}\!\left(\big\| \hat{u}_{t+1} - u_{t+1} \big\|_2^2\right).
\end{align}

\subsubsection{RND baseline} \label{sec:intrinsic_rnd}
RND~\cite{burda_exploration_2018} uses a fixed random target \(f\) and a predictor \(p\) over latent features \(x_t\); in our case we set \(x_t = u_t = [h_t; s_t]\):

\vspace{-15pt} 
\begin{align}
\mathcal{L}^{\text{RND}} &= \| p(x_t) - f(x_t) \|_2^2, \\
r^{\text{int}}_{\text{RND},t} &= \mathrm{Norm}\!\left(\| p(x_t) - f(x_t) \|_2^2\right).
\end{align}

All intrinsic modules (disagreement, ICM, RND) receive the \emph{same} latent features and share replay, schedules, and budgets. 
Normalization uses running statistics over the unnormalized scalar intrinsic signal $g_t$:
\(
\mathrm{Norm}(g_t)=\frac{g_t-\mu_t}{\sigma_t+\epsilon},\ 
\mu_t\!\leftarrow\!(1\!-\!\rho)\mu_{t-1}\!+\!\rho g_t,\ 
\sigma_t^2\!\leftarrow\!(1\!-\!\rho)\sigma_{t-1}^2\!+\!\rho(g_t-\mu_t)^2,
\)
where $\rho \in (0,1)$ is the exponential moving-average rate.

\subsection{Actor–Critic Learning with Imagined Returns}
\label{sec:imagined}

We train an actor--critic in imagination using imagined rollouts from the RSSM.

\paragraph*{Pretraining (exploration policy)}
During reward-free pretraining, we train only an exploration policy $\pi_{\theta}^{\text{expl}}$ and value $V_{\psi}^{\text{expl}}$
to maximize intrinsic return:
\begin{equation}
J_{\text{explore}}
  = \mathbb{E}\left[\sum_{\tau=t}^{t+H-1} \gamma^{\tau-t} r^{\text{int}}_{\tau}\right].
\label{eq:explore_objective}
\end{equation}

where $r^{\text{int}}$ is computed by the selected intrinsic module (LD/ICM/RND).

\paragraph*{Fine-tuning (task policy)}
During few-shot fine-tuning, we introduce a separate task policy $\pi_{\theta}^{\text{task}}$ (initialized from $\pi_{\theta}^{\text{expl}}$) and train it to maximize extrinsic return:
\begin{equation}
J_{\text{task}}
  = \mathbb{E}\left[\sum_{\tau=t}^{t+H-1} \gamma^{\tau-t} r^{\text{ext}}_{\tau}\right].
\label{eq:task_objective}
\end{equation}

Where H is the imagination rollout horizon. When imagined rollouts are used for fine-tuning, $r^{\text{ext}}$ is provided by an extrinsic reward head trained only in Phase~2.





\begin{figure*}[t]
  \centering

  \begin{subfigure}[t]{0.69\textwidth}
    \centering
    \includegraphics[scale=0.5]{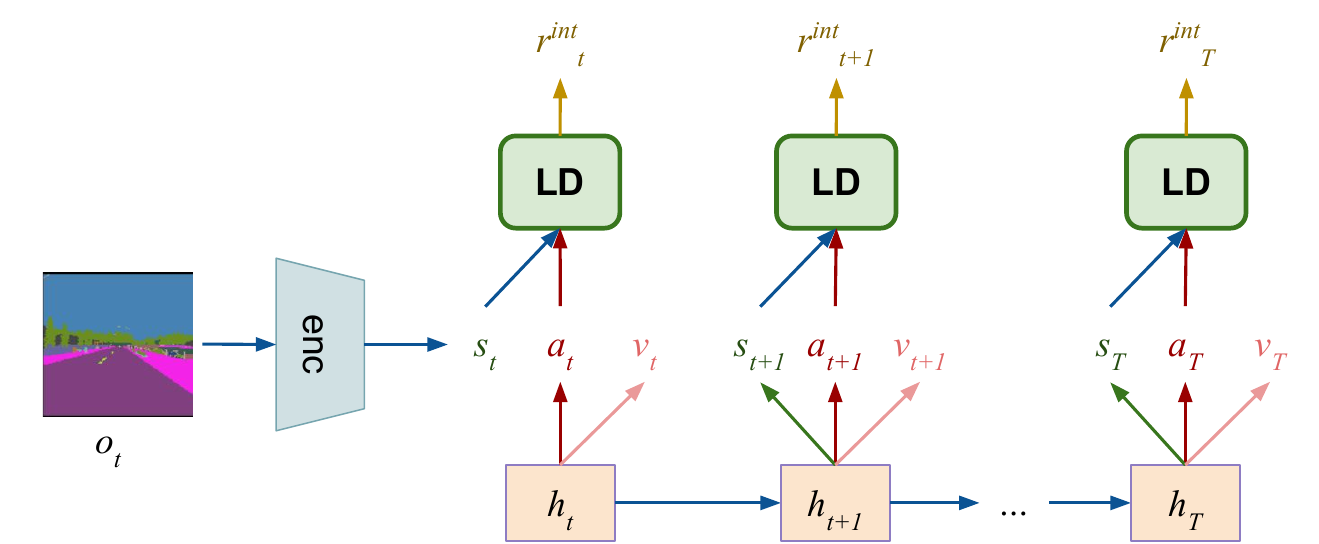}
    \captionsetup{font=footnotesize} 
    \caption{Overview of the InDRiVE Actor Critic Policy.}
    \label{fig:method_overview_a}
  \end{subfigure}
  \hfill
  \begin{subfigure}[t]{0.3\textwidth}
    \centering
    \includegraphics[scale=0.45]{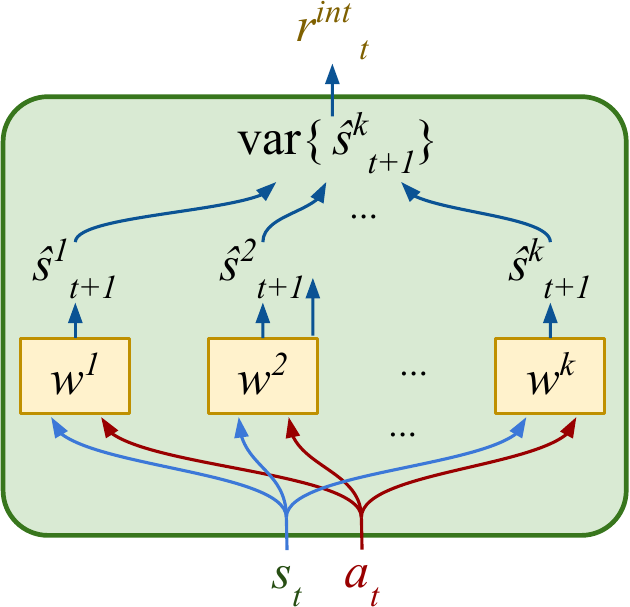}
    \captionsetup{font=footnotesize} 
    \caption{Latent Disagreement (LD) Reward}
    \label{fig:method_overview_b}
  \end{subfigure}

  \captionsetup{font=footnotesize} 
  \caption{%
    Overview of InDRiVE. 
    (a) An actor critic policy architecture incorporating latent disagreement for exploration. LD is Latent Disagreement in (b). Raw images are encoded into a stochastic latent $s_t$, which is combined with deterministic hidden state $h_t$ to maintain temporal context. The actor--critic policy then outputs an action $a_t$ based on $u_t = [s_t, h_t]$. 
    (b) An \textit{ensemble} of forward 
    models predicts potential next states $\hat{s}_{t+1}^{\,k}$ for the same $(s_t, a_t)$. The variance among these predictions yields a latent-disagreement (intrinsic) reward, which, encourages the policy to explore.}
  \label{fig:method_overview}
\end{figure*}

The resulting actor–critic architecture and latent-disagreement module are illustrated in Fig.~\ref{fig:method_overview}.

\subsection{Two-Phase Protocol: Zero-Shot and Fine-Tuning}
\label{sec:phases}
The overall pipeline separates deployment into two phases:

\paragraph*{Intrinsic pretraining (reward-free)}
Collect trajectories in CARLA using the exploration policy.
Update the world model by minimizing $\mathcal{L}_\text{model}^\text{pre}$ (Eq.~\ref{eq:wm_pre}) and update the
intrinsic module (disagreement / ICM / RND). Train the exploration actor--critic in imagination to maximize $J_{\text{explore}}$.
No task reward is computed, stored, or used during this phase.

\paragraph*{Phase~1 — Zero-shot transfer (no updates)}
Freeze all parameters and evaluate the pretrained \emph{exploration policy} in an unseen town and routes without any additional data collection or gradient steps. Downstream metrics are computed only for evaluation.

\paragraph*{Phase~2 — Fine-tuning (few-shot)}
Introduce a task-specific extrinsic reward $r_t^{\text{ext}}$ and train a task policy for a small interaction budget.
Unless stated otherwise, keep the dynamics/decoder frozen and update only the task policy/value and the extrinsic reward head
(imagined rollouts require $\hat r_t^{\text{ext}}$).



\section{Experimental Setup}
\label{sec:experimental-setup}

\noindent

This section details the experimental framework for assessing our proposed approach. We begin by introducing the CARLA simulation environment and the tasks under consideration, followed by the two-phase training procedure. We then describe the baseline methods, hyperparameter configurations, and the metrics used for evaluation.


\subsection{Environment Setup}
\label{sec:env-setup}

\noindent

We conducted experiments in the CARLA simulator using two towns: \texttt{Town01} (a small town with a river and bridges) for training and \texttt{Town02} (a small town with a mixture of residential and commercial buildings) for testing.

\noindent

\paragraph{Observations}

At each step (10\,Hz), the agent receives a $128\times 128$ front-view semantic segmentation image from CARLA. We represent the segmentation as a 3-channel RGB-coded mask (CARLA palette) and normalize pixel values to $[0,1]$. The policy input consists only of this front-view semantic-segmentation image; we do not provide high-level route commands, waypoint coordinates, or privileged simulator state to the policy—route waypoints are used only for reward computation and evaluation metrics.

\paragraph{Actions}
Actions are discrete low-level controls applied at 10\,Hz.
We use $\mathcal{A}=\mathcal{A}_{\text{steer}}\times \mathcal{A}_{\text{throttle}}$ with
$\mathcal{A}_{\text{throttle}}=\{0.0,0.2,0.3\}$ and $\mathcal{A}_{\text{steer}}=\{-0.3,0.0,0.3\}$.
We fix brake to $0.0$ to keep the action space minimal; the vehicle slows via coasting and termination events.

\paragraph{Simulator and control}
Experiments use CARLA~\texttt{0.9.15} in synchronous mode with fixed $\Delta t{=}0.1$\,s (10\,Hz control; cameras 10\,Hz). We set nearby vehicle density in $\{5,10,20\}$ within 150\,m of ego via the TrafficManager (seed $s_{\text{TM}}$). We fix the global RNG seeds for reproducibility. During evaluation, we average results over multiple episodes with randomized TrafficManager and ego-spawn seeds; the episode counts and seed sets are specified in Sec \ref{metrics}. All non-ego vehicles are controlled by CARLA TrafficManager autopilot with default parameters.


\subsection{Pretraining with Intrinsic Reward}

During pretraining, we train the DreamerV3 world model and an exploration policy using only the intrinsic reward from Sec.~\ref{sec:intrinsic}. InDRiVE and the parity baselines (ICM, RND) are pretrained for $5\times 10^5$ environment steps in \texttt{Town01}.
We prefill the replay buffer with $\mathcal{T}_{\text{prefill}}=1000$ random steps before learning begins. Episodes terminate using the conditions in Sec.~\ref{sec:termination}; after termination, we reset and respawn the ego vehicle at a random spawn point in \texttt{Town01}. To encourage diversity, we resample traffic density from $\{5,10,20\}$ vehicles (within 150\,m) every 10{,}000 steps. We do not constrain exploration to fixed routes during pretraining.


\subsection{Fine-tuning on Task-specific Reward}
After intrinsic pretraining, we perform few-shot adaptation for $10$k environment steps using task-specific extrinsic rewards on the fixed routes in \texttt{Town01} (Fig.~\ref{fig:routes_town01}). During few-shot fine-tuning we freeze the pretrained world model
(encoder, RSSM dynamics, decoder, and continuation head) and freeze the intrinsic module. We train only (i) the task actor--critic networks and (ii) an extrinsic reward head $p_{\phi_r}(r_t^{ext}\mid h_t,s_t)$ used to provide rewards for imagined rollouts.

\subsubsection{Downstream Tasks}
\label{sec:tasks-scenarios}

We consider two representative driving tasks to demonstrate zero-shot and few-shot performance:

\begin{itemize}
    \item \textbf{Lane Following (LF):} The agent must maintain its lane position while traveling at a safe speed.
    \item \textbf{Collision Avoidance (CA):} The agent must avoid colliding with other vehicles and obstacles in real-time traffic scenarios.
\end{itemize}

We use different reward weights for LF and CA (Table~\ref{tab:reward_weights}) to reflect different objectives.

\subsubsection{Reward Design for Fine-tuning}

\begin{table}[t]
\centering
\captionsetup{font=footnotesize} 
\caption{Reward weights used for fine-tuning Lane Following (LF) and Collision Avoidance (CA).}

\label{tab:reward_weights}
\begin{tabular}{lcc}
\toprule
Parameter / weight & LF & CA \\
\midrule
Waypoint progress weight $w_{\text{waypt}}$     & 1.0   & 0.5 \\
Target speed $v_{\star}$ [m/s]                  & 5.0   & 3.0 \\
Speed reward scale $\alpha_{\text{speed}}$      & 0.05  & 0.03 \\
Lane-keeping penalty weight $w_{\text{lane}}$   & 2.0   & 1.0 \\
Collision penalty weight $w_{\text{coll}}$      & 30.0  & 80.0 \\
Destination bonus $w_{\text{dest}}$             & 100   & 100 \\
Per-step time penalty $w_{\text{time}}$         & 0.01  & 0.01 \\
\bottomrule
\end{tabular}
\end{table}


\noindent At each control step, the extrinsic reward is computed as the sum of the following components
(the same components are used for Lane Following and Collision Avoidance; only the weights differ as shown
in Table~\ref{tab:reward_weights}):

\begin{itemize}[leftmargin=*,noitemsep,topsep=2pt,parsep=0pt,partopsep=0pt]
  \item \textbf{Waypoint progress.} If the agent advances at least one route waypoint since the previous step,
  add a fixed progress bonus; otherwise add nothing.

  \item \textbf{Speed tracking.} Use the heading of the next route waypoint to define the forward direction.
  Reward maintaining the task’s target forward speed by subtracting the absolute forward-speed error.
  Penalize sideways (lateral) motion as well, with the lateral-speed penalty capped at 0.5\,m/s and weighted twice
  as strongly as the forward-speed error. Scale the resulting term using the speed scale from Table~\ref{tab:reward_weights}.

  \item \textbf{Lane keeping.} Compute the lateral offset from the lane centerline and the lane width from the route waypoints.
  Apply no penalty while the offset stays within a 20\% lane-width tolerance band.
  Once outside the tolerance band, apply a quadratic penalty that increases with offset and saturates at its maximum
  once the offset reaches half of the lane width. Multiply by the lane-keeping weight.

  \item \textbf{Collision.} If a collision is detected, apply a negative penalty proportional to the vehicle’s current speed magnitude,
  scaled by the collision weight.

  \item \textbf{Destination.} When the destination is reached, add a terminal success bonus.

  \item \textbf{Time.} Apply a small constant negative reward every step to discourage unnecessarily long episodes.
\end{itemize}


\subsection{Termination Condition for Training and Evaluation} \label{sec:termination}

Episodes terminate upon any of the following events:
\begin{itemize}
    \item Collision: The agent collides with another vehicle, pedestrian, or static obstacle.
    \item Wrong Direction: The agent drives in the opposite direction of the intended lane.
    \item Off-Road Driving: The agent leaves the drivable area.
    \item Vehicle Stall: The agent's velocity falls below a minimal threshold (i.e, 1~km/h) for an extended period (1~minute).
    \item Destination reached: The agent reaches the destination (terminal waypoint) or completes the loop for closed routes.
    \item Time limit: The episode terminates after 1000 steps if the destination has not been reached.
    \item Lane violation: We use CARLA’s lane invasion sensor to detect lane-marking crossings. We count any lane invasion as a failure and terminate the episode at the first invasion event.

\end{itemize}


\subsection{Evaluation Routes}
\label{sec:eval-routes}

\begin{figure*}[t]
  \centering
  \begin{subfigure}[t]{0.48\textwidth}
    \centering
    \includegraphics[width=\linewidth]{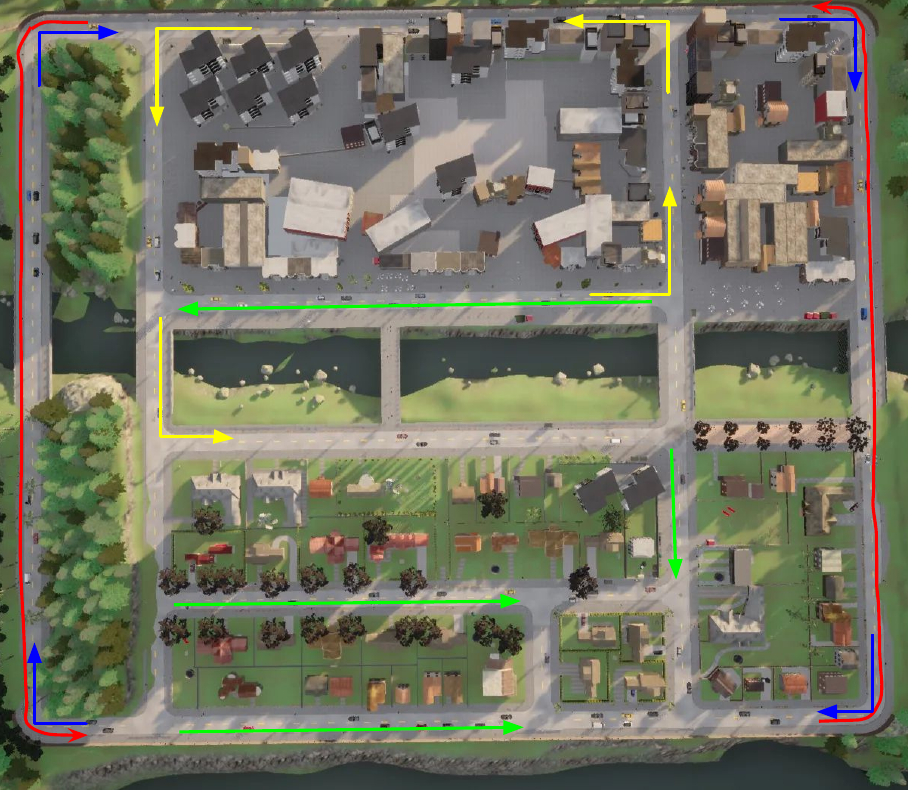}
    \caption{Routes Town01}
    \label{fig:routes_town01}
  \end{subfigure}
  \hfill
  \begin{subfigure}[t]{0.44\textwidth}
    \centering
    \includegraphics[width=\linewidth]{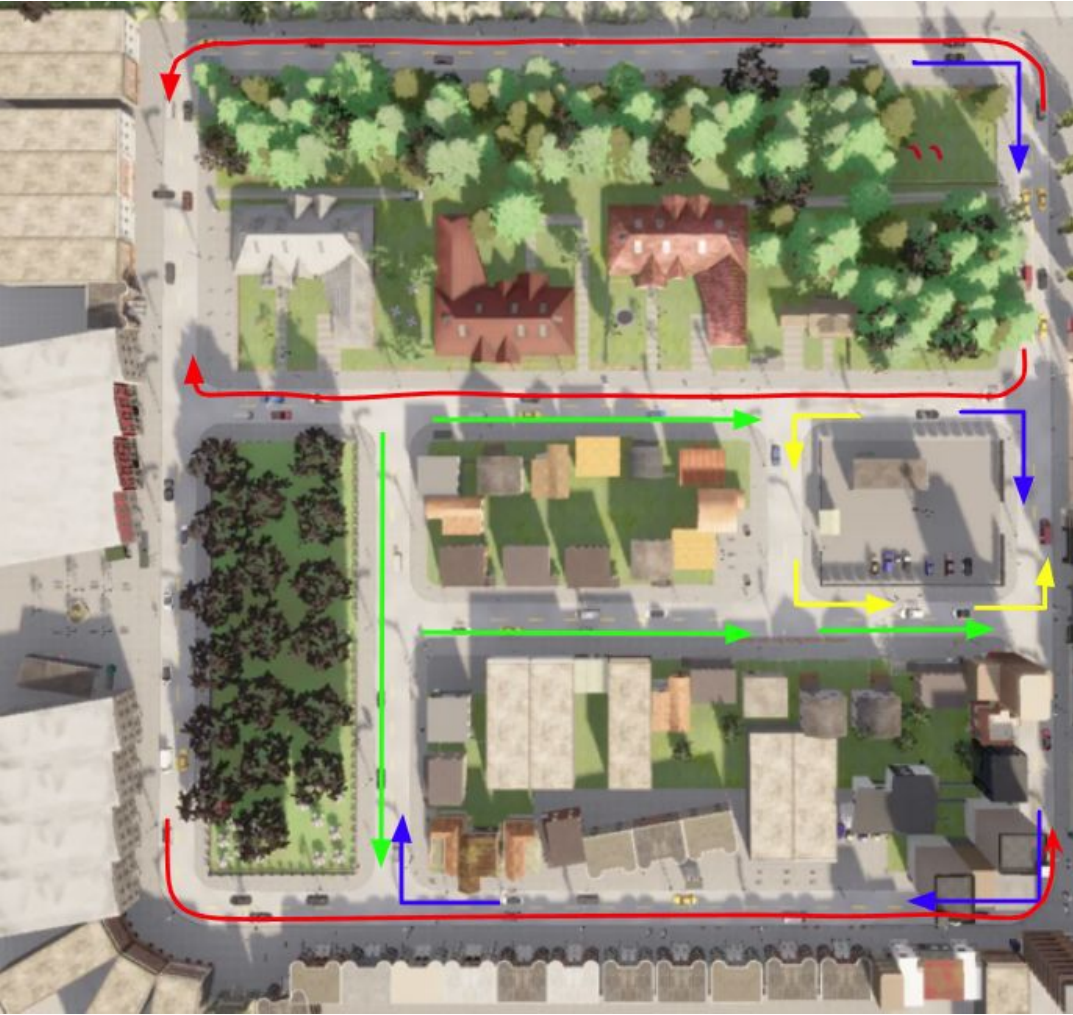}
    \caption{Routes Town02}
    \label{fig:routes_town02}
  \end{subfigure}

  \captionsetup{font=footnotesize}
  \caption{Evaluation routes in CARLA for transfer tests. Top‑down maps for Town01 (seen) (a) and Town02 (unseen) (b). Colored trajectories denote the four fixed routes: Straight (green), Right‑Turn loop (blue), Left‑Turn loop (yellow), and Two‑Turn loop (red); arrows indicate travel direction. Each route is evaluated under traffic densities {5, 10, 20} vehicles within 150 m and averaged across seeds.}
  \vspace{-10pt}
  \label{fig:routes}
\end{figure*}

We use four color-coded routes (Fig.~\ref{fig:routes}) that isolate core driving skills. Arrows indicate the fixed travel direction.

\begin{itemize}
    \item \textbf{Red — Two-Turn Route:} Closed loop with a mix of left and right turns across several intersections. Stresses long-horizon planning.
    \item \textbf{Yellow — Left-Turn Route:} Compact loop consisting exclusively of left turns. 
    \item \textbf{Blue — Right-Turn Route:} Compact loop consisting exclusively of right turns.
    \item \textbf{Green — Straight Route:} Point-to-point straight corridor with no turns. 
\end{itemize}

\paragraph{Protocol.}
For each town (\texttt{Town01}, \texttt{Town02}), we instantiate the four routes with CARLA waypoints and evaluate at traffic densities $\{5,10,20\}$ vehicles within 150\,m, matching training. At episode start, the ego is spawned at the start waypoint on the selected route. Success is recorded when the agent completes one full loop (red/yellow/blue) or reaches the terminal waypoint (green) within the step budget and without failures from Sec.~\ref{sec:termination}. Each route is run over multiple random traffic seeds, and metrics from Sec.~\ref{metrics} are reported per route and averaged over seeds. We report per-route SR as the uniform average over traffic densities {5,10,20}, then averaged over training seeds.

The evaluated policies are not conditioned on route commands or a goal; the ego vehicle is spawned at a route start pose, and the route waypoints are used only for evaluation/progress computation and wrong-direction checks, not as policy inputs. Therefore, SR in Sec \ref{metrics} should be interpreted as robust low-level driving and intersection-handling along standardized start-to-end corridors, not as goal-conditioned navigation or global planning.




\subsection{Baseline Methods}
\label{sec:baselines}

We compare \textbf{InDRiVE} against three baselines under matched interaction budgets and identical backbones (encoder, latent size, optimizer, observation) and identical evaluation protocols.

\paragraph{Extrinsic-only baselines}
\begin{enumerate}
  \item \textbf{DreamerV3 (task-specific).} World model and policy are trained from scratch using only task rewards.
  For fairness, DreamerV3 is trained for the same total interaction budget as InDRiVE (500k + 10k = 510k steps) on the task reward in Town01. We train separate DreamerV3 agents for Lane Following and Collision Avoidance using the corresponding reward weights in Table \ref{tab:reward_weights}.

\end{enumerate}

\paragraph{Intrinsic-pretrain parity baselines.}
To isolate the contribution of the intrinsic signal, we implement curiosity modules in the \emph{Dreamer latent space} and keep all other components identical:
\begin{enumerate}
  \item \textbf{DreamerV3+ICM}, implemented exactly as Sec \ref{sec:intrinsic_icm}, operating on $\mathbf{u}_t = [\mathbf{h}_t;\,\mathbf{s}_t]$.
  \item \textbf{DreamerV3+RND}, implemented exactly as Sec \ref{sec:intrinsic_rnd}, operating on $\mathbf{u}_t = [\mathbf{h}_t;\,\mathbf{s}_t]$.
\end{enumerate}

\paragraph{Training protocol.}
All intrinsic-pretrain methods (InDRiVE, ICM, RND) are trained for \textbf{500k} environment steps using intrinsic reward only; the world model and actor--critic are optimized in latent space. Zero-shot evaluation freezes the world model (and the intrinsic module). Few-shot fine-tuning then updates the task policy/value networks and trains the extrinsic
reward head for 10k environment steps using extrinsic rewards, while keeping the pretrained
world model and intrinsic module frozen.

\vspace{4pt}

\begin{table}[htbp]
    \centering
    \captionsetup{font=footnotesize} 
    \caption{Key hyperparameters for intrinsic pretraining and few-shot fine-tuning. All methods share the same encoder, latent size, optimizer, and observation.}
    \label{tab:hyperparams}
    \resizebox{\columnwidth}{!}{%
    \begin{tabular}{lcc}
    \toprule
    \textbf{Hyperparameter} & \textbf{Intrinsic Pretrain} & \textbf{Few-shot Fine-tuning} \\
    \midrule
    Ensemble size ($K$; InDRiVE)  & 8                  & 8 (frozen) \\
    Batch size                    & 64                 & 64 \\
    Replay buffer size            & $10^{5}$           & $10^{5}$ \\
    Discount factor ($\gamma$)    & 0.99               & 0.99 \\
    Intrinsic weight ($\alpha$)   & 1.0 (intrinsic-only) & 0.0 (extrinsic-only) \\
    Randomize cadence / rollout   & 10k / 1k steps     & -- \\
    \textbf{Env. steps (per phase)} & 500k     & 10k \\
    Normalization epsilon $\epsilon$ & 1.00 & 1.00 \\
    Imagination horizon $H$ & 15 & 15 \\
    \bottomrule
    
\vspace{-22pt} 
    \end{tabular}%
    }
\end{table}

\subsection{Evaluation Metrics}\label{metrics}
For each town–route–method–traffic density configuration, we run N = 50 evaluation episodes. Traffic density is fixed per configuration; TrafficManager seeds are sampled from a fixed set of 3 seeds and ego-spawn seeds are sampled from a fixed set of 5 seeds.
An episode is labeled \textbf{successful} if the agent reaches the destination within the step budget and the logger reports
no collision, no lane violation, no wrong-direction driving, and no off-road event during the episode.
The \textbf{Success Rate (SR)} is the percentage of successful episodes. We present a separate Failure Rate = $\mathrm{FR}=100-\mathrm{SR}$ by construction.

\section{Results}
\label{sec:results}

\begin{table*}[t]
    \centering
    \captionsetup{font=footnotesize} 
    \caption{Zero-shot transfer on Town01 (seen) and Town02 (unseen) across the four evaluation routes (Fig. \ref{fig:routes}). Policies are frozen (no updates). Values are SR/FR (\%) computed under the evaluation protocol in Sec. \ref{metrics}}
    \label{tab:zeroshot_town01_town02}
    \footnotesize
    
    \setlength{\tabcolsep}{2pt}
    \renewcommand{\arraystretch}{1.1}

    \begin{tabular}{l*{4}{cccc}}
        \toprule
        & \multicolumn{4}{c}{Straight}
        & \multicolumn{4}{c}{Right Turn}
        & \multicolumn{4}{c}{Left Turn}
        & \multicolumn{4}{c}{Two Turn} \\
        \cmidrule(lr){2-5}\cmidrule(lr){6-9}\cmidrule(lr){10-13}\cmidrule(lr){14-17}
        & \multicolumn{2}{c}{Town01} & \multicolumn{2}{c}{Town02}
        & \multicolumn{2}{c}{Town01} & \multicolumn{2}{c}{Town02}
        & \multicolumn{2}{c}{Town01} & \multicolumn{2}{c}{Town02}
        & \multicolumn{2}{c}{Town01} & \multicolumn{2}{c}{Town02} \\
        \cmidrule(lr){2-3}\cmidrule(lr){4-5}
        \cmidrule(lr){6-7}\cmidrule(lr){8-9}
        \cmidrule(lr){10-11}\cmidrule(lr){12-13}
        \cmidrule(lr){14-15}\cmidrule(lr){16-17}
        Method
        & \thead{SR ($\uparrow$)}
        & \thead{FR ($\downarrow$)}
        & \thead{SR ($\uparrow$)}
        & \thead{FR ($\downarrow$)}
        & \thead{SR ($\uparrow$)}
        & \thead{FR ($\downarrow$)}
        & \thead{SR ($\uparrow$)}
        & \thead{FR ($\downarrow$)}
        & \thead{SR ($\uparrow$)}
        & \thead{FR ($\downarrow$)}
        & \thead{SR ($\uparrow$)}
        & \thead{FR ($\downarrow$)}
        & \thead{SR ($\uparrow$)}
        & \thead{FR ($\downarrow$)}
        & \thead{SR ($\uparrow$)}
        & \thead{FR ($\downarrow$)} \\
        \midrule
        InDRiVE
        & \textbf{76.50}  & \textbf{23.50}
        & \textbf{75.00}   & \textbf{25.00}
        & \textbf{85.50}  & \textbf{14.50}
        & \textbf{82.10}  & \textbf{17.90}
        & \textbf{75.40}  & \textbf{24.60}
        & \textbf{70.90}  & \textbf{29.10}
        & \textbf{65.20}  & \textbf{34.80}
        & \textbf{64.78} & \textbf{35.22} \\
        ICM
        & 68.70  & 31.30
        & 45.20  & 54.80
        & 61.20  & 38.80
        & 44.90  & 55.10
        & 63.00  & 37.00
        & 55.10  & 44.90
        & 57.40  & 42.60
        & 49.23 & 50.77 \\
        RND
        & 71.50  & 28.50
        & 62.90  & 37.10
        & 79.20  & 20.80
        & 60.22 & 39.80
        & 65.50  & 34.50
        & 53.10  & 46.90
        & 58.10  & 41.90
        & 47.10  & 52.90 \\
        \bottomrule

    \end{tabular}
\end{table*}

\begin{table*}[t]
    \centering
    \captionsetup{font=footnotesize} 
    \caption{Few-shot adaptation (Lane Following) after 10k extrinsic-reward steps. SR/FR (\%) on Town01 (seen) and Town02 (unseen) for each route (Fig. \ref{fig:routes}). Fine-tuning updates only the task policy/value (and reward head); the pretrained world model and intrinsic module remain frozen (Sec. \ref{sec:baselines})}
    \label{tab:fewshot_lanefollowing_town01_town02}
    \footnotesize
    
    \setlength{\tabcolsep}{2pt}
    \renewcommand{\arraystretch}{1.1}

    \begin{tabular}{l*{4}{cccc}}
        \toprule
        & \multicolumn{4}{c}{Straight}
        & \multicolumn{4}{c}{Right Turn}
        & \multicolumn{4}{c}{Left Turn}
        & \multicolumn{4}{c}{Two Turn} \\
        \cmidrule(lr){2-5}\cmidrule(lr){6-9}\cmidrule(lr){10-13}\cmidrule(lr){14-17}
        & \multicolumn{2}{c}{Town01}
        & \multicolumn{2}{c}{Town02}
        & \multicolumn{2}{c}{Town01}
        & \multicolumn{2}{c}{Town02}
        & \multicolumn{2}{c}{Town01}
        & \multicolumn{2}{c}{Town02}
        & \multicolumn{2}{c}{Town01}
        & \multicolumn{2}{c}{Town02} \\
        \cmidrule(lr){2-3}\cmidrule(lr){4-5}
        \cmidrule(lr){6-7}\cmidrule(lr){8-9}
        \cmidrule(lr){10-11}\cmidrule(lr){12-13}
        \cmidrule(lr){14-15}\cmidrule(lr){16-17}
        Method
        & \thead{SR ($\uparrow$)}
        & \thead{FR ($\downarrow$)}
        & \thead{SR ($\uparrow$)}
        & \thead{FR ($\downarrow$)}
        & \thead{SR ($\uparrow$)}
        & \thead{FR ($\downarrow$)}
        & \thead{SR ($\uparrow$)}
        & \thead{FR ($\downarrow$)}
        & \thead{SR ($\uparrow$)}
        & \thead{FR ($\downarrow$)}
        & \thead{SR ($\uparrow$)}
        & \thead{FR ($\downarrow$)}
        & \thead{SR ($\uparrow$)}
        & \thead{FR ($\downarrow$)}
        & \thead{SR ($\uparrow$)}
        & \thead{FR ($\downarrow$)} \\
        \midrule
        InDRiVE
        & \textbf{100.00}   & \textbf{0.00}
        & 99.00             & 1.00
        & 97.45          & 2.55
        & \textbf{92.18} & \textbf{7.82}
        & 98.20           & 1.80
        & \textbf{95.10}  & \textbf{4.90}
        & 94.70           & 5.30
        & \textbf{88.41} & \textbf{11.59} \\
        ICM
        & 89.70   & 10.30
        & 89.47  & 10.53
        & 83.07  & 16.93
        & 80.21  & 19.79
        & 80.09  & 19.91
        & 78.47  & 21.53
        & 74.02  & 25.98
        & 70.19  & 29.81 \\
        RND
        & \textbf{100.00} & \textbf{0.00}
        & \textbf{100.00} & \textbf{0.00}
        & 98.59 & 1.41
        & 98.04 & 1.96
        & 94.23 & 5.79
        & 91.88 & 8.12
        & 90.00    & 10.00
        & 90.00   & 10.00 \\
        DreamerV3
        & \textbf{100.00} & \textbf{0.00}
        & 94.21 & 5.79
        & \textbf{100.00} & \textbf{0.00}
        & 90.02 & 9.98
        & \textbf{100.00} & \textbf{0.00}
        & 88.43 & 11.57
        & \textbf{100.00} & \textbf{0.00}
        & 83.40 & 16.60 \\
        \bottomrule

    \end{tabular}
\end{table*}

\begin{table*}[t]
    \centering
    \captionsetup{font=footnotesize} 
    \caption{Few-shot adaptation (Collision Avoidance) after 10k extrinsic-reward steps. SR/FR (\%) on Town01 (seen) and Town02 (unseen) across the evaluation routes (Fig. \ref{fig:routes}). Fine-tuning protocol matches Table IV (Sec. \ref{sec:baselines})}
    \label{tab:fewshot_collision_town01_town02}
    \footnotesize
    \setlength{\tabcolsep}{2pt}
    \renewcommand{\arraystretch}{1.1}

    \begin{tabular}{l*{4}{cccc}}
        \toprule
        & \multicolumn{4}{c}{Straight}
        & \multicolumn{4}{c}{Right Turn}
        & \multicolumn{4}{c}{Left Turn}
        & \multicolumn{4}{c}{Two Turn} \\
        \cmidrule(lr){2-5}\cmidrule(lr){6-9}\cmidrule(lr){10-13}\cmidrule(lr){14-17}
        & \multicolumn{2}{c}{Town01}
        & \multicolumn{2}{c}{Town02}
        & \multicolumn{2}{c}{Town01}
        & \multicolumn{2}{c}{Town02}
        & \multicolumn{2}{c}{Town01}
        & \multicolumn{2}{c}{Town02}
        & \multicolumn{2}{c}{Town01}
        & \multicolumn{2}{c}{Town02} \\
        \cmidrule(lr){2-3}\cmidrule(lr){4-5}
        \cmidrule(lr){6-7}\cmidrule(lr){8-9}
        \cmidrule(lr){10-11}\cmidrule(lr){12-13}
        \cmidrule(lr){14-15}\cmidrule(lr){16-17}
        Method
        & \thead{SR ($\uparrow$)}
        & \thead{FR ($\downarrow$)}
        & \thead{SR ($\uparrow$)}
        & \thead{FR ($\downarrow$)}
        & \thead{SR ($\uparrow$)}
        & \thead{FR ($\downarrow$)}
        & \thead{SR ($\uparrow$)}
        & \thead{FR ($\downarrow$)}
        & \thead{SR ($\uparrow$)}
        & \thead{FR ($\downarrow$)}
        & \thead{SR ($\uparrow$)}
        & \thead{FR ($\downarrow$)}
        & \thead{SR ($\uparrow$)}
        & \thead{FR ($\downarrow$)}
        & \thead{SR ($\uparrow$)}
        & \thead{FR ($\downarrow$)} \\
        \midrule
        InDRiVE
        & \textbf{100.00}   & \textbf{0.00}
        & \textbf{100.00}   & \textbf{0.00}
        & \textbf{100.00}   & \textbf{0.00}
        & \textbf{100.00}   & \textbf{0.00}
        & \textbf{95.74} & \textbf{4.26}
        & \textbf{93.57} & \textbf{6.43}
        & \textbf{89.99} & \textbf{10.01}
        & \textbf{85.55} & \textbf{14.45} \\
        ICM
        & 98.02 & 1.98
        & 89.47 & 10.53
        & 95.47 & 4.53
        & 83.07 & 16.93
        & 90.00    & 10.00
        & 89.36 & 10.64
        & 88.32 & 11.68
        & 78.21 & 21.79 \\
        RND
        & \textbf{100.00} & \textbf{0.00}
        & 95.20         & 4.80
        & \textbf{100.00} & \textbf{0.00}
        & 91.23        & 8.77
        & 92.54        & 7.45
        & 88.45        & 11.55
        & 82.00           & 18.00
        & 80.29        & 19.71 \\
        DreamerV3
        & \textbf{100.00} & \textbf{0.00}
        & 92.71        & 7.29
        & 98.30         & 1.70
        & 97.29        & 2.71
        & 88.68        & 11.32
        & 85.52        & 14.48
        & 81.48        & 18.52
        & 78.02        & 21.98 \\
        \bottomrule

    \end{tabular}
\end{table*}

\begin{table}[t]
\centering
\captionsetup{font=footnotesize} 
\caption{Average SR (\%) and generalization gap across towns. Averages are uniform over the four routes.
Gap is Town01 SR minus Town02 SR. DreamerV3 is trained task-specifically (not reward-free).}
\label{tab:gap}
\begin{tabular}{lcccc}
\toprule
\textbf{Setting} & \textbf{Method} & \textbf{Town01} & \textbf{Town02} & \textbf{Gap ($\downarrow$)} \\
\midrule
\multirow{3}{*}{Zero-shot (Table III)} 
& InDRiVE & \textbf{75.65} & \textbf{73.20} & \textbf{2.45} \\
& ICM     & 62.58 & 48.61 & 13.97 \\
& RND     & 68.58 & 55.83 & 12.75 \\
\midrule
\multirow{4}{*}{Few-shot LF (Table IV)} 
& InDRiVE  & 97.59 & 93.67 & 3.92 \\
& ICM      & 81.72 & 79.59 & 2.13 \\
& RND      & 95.71 & \textbf{94.98} & \textbf{0.73} \\
& DreamerV3& \textbf{100.00} & 89.01 & 10.99 \\
\midrule
\multirow{4}{*}{Few-shot CA (Table V)} 
& InDRiVE  & \textbf{96.43} & \textbf{94.78} & \textbf{1.65} \\
& ICM      & 92.95 & 85.03 & 7.92 \\
& RND      & 93.64 & 88.79 & 4.85 \\
& DreamerV3& 92.12 & 88.38 & 3.74 \\
\bottomrule
\vspace{-25pt} 
\end{tabular}
\end{table}

We pretrained InDRiVE and intrinsic baselines (ICM, RND) for $5\times10^5$ reward-free steps in \texttt{Town01}, then evaluate (i) \emph{zero-shot} on the pretrained policy without any adaptation, and (ii) \emph{few-shot} after $10$k steps of task-reward fine-tuning in \texttt{Town01}. We repeated training with 5 random seeds and report mean performance; each seed is evaluated with N = 50 episodes per (town, route, traffic density), using TrafficManager seeds sampled from a fixed set of 3 seeds and spawn seeds sampled from a fixed set of 5 seeds (seed pairs may repeat).

\subsection{Zero-shot transfer across towns}
Table~\ref{tab:zeroshot_town01_town02} reports zero-shot success rate (SR). InDRiVE is consistently strongest across routes and transfers more reliably to the unseen \texttt{Town02}. Averaged over routes, InDRiVE achieves $75.7\%$ SR in \texttt{Town01} and $73.2\%$ SR in \texttt{Town02}, a drop of only $\approx2.5$ points, whereas ICM and RND degrade by $\approx14.0$ and $\approx12.7$ points, respectively.
Gains are most pronounced on turn-heavy routes under distribution shift (e.g., Right-Turn in \texttt{Town02}: $82.1\%$ SR for InDRiVE vs.\ $44.9\%$ for ICM and $60.2\%$ for RND), indicating that disagreement-driven exploration yields a more transferable world model.

\subsection{Few-shot adaptation}
Tables~\ref{tab:fewshot_lanefollowing_town01_town02} and~\ref{tab:fewshot_collision_town01_town02} summarize few-shot adaptation after only $10$k extrinsic-reward steps.

\paragraph{Lane following.}
After fine-tuning, InDRiVE reaches high SR on \texttt{Town01} (94.7--100\% across routes) and maintains strong performance on \texttt{Town02} (88.4--99\%). Relative to DreamerV3 trained from scratch with the matched total interaction budget, InDRiVE improves generalization on \texttt{Town02} by roughly 2--7 points depending on the route, with the largest gains on left-/multi-turn routes. ICM remains the weakest baseline after adaptation (typically 70--89\% SR on \texttt{Town02}), suggesting that prediction-error curiosity is less effective for building transferable driving dynamics in this setting.

\paragraph{Collision avoidance.}
InDRiVE is strongest and most robust for safety-critical behavior. It achieves perfect SR on Straight and Right-Turn routes in both towns and remains best on the more challenging turn-heavy routes (e.g., Two-Turn in \texttt{Town02}: $85.6\%$ SR for InDRiVE vs.\ $80.3\%$ RND, $78.2\%$ ICM, $78.0\%$ DreamerV3). Averaged over routes, InDRiVE achieves $96.4\%$ SR on \texttt{Town01} and $94.8\%$ SR on \texttt{Town02}, with the smallest degradation under town shift among all methods.

Table~\ref{tab:gap} highlights that disagreement pretraining yields the smallest cross-town degradation in the zero-shot setting
and the most robust few-shot collision-avoidance behavior under town shift, while lane-following transfer is competitive with RND.

Overall, these results show that reward-free disagreement pretraining produces a reusable driving world model: it improves zero-shot generalization and enables rapid few-shot specialization, particularly for collision avoidance under distribution shift.

\section{Conclusion and Future Work}

We introduced InDRiVE, a reward-free pretraining framework for autonomous driving, in which the world model and exploration policy are trained solely from ensemble disagreement signals, followed by few-shot specialization with limited task reward. InDRiVE achieves comparable lane-following performance to task-trained DreamerV3 with better generalization to Town02, and clearly higher success rates and fewer failures for collision avoidance under town shift. Its latent representation transfers effectively to both familiar (\texttt{Town01}) and unfamiliar (\texttt{Town02}) settings, enabling zero-shot or few-shot adaptation to tasks like lane-following and collision avoidance. These findings highlight the benefits of purely intrinsic exploration in uncovering robust driving policies and underscore the potential for reducing dependence on manual reward design. Our evaluation is limited to CARLA Town01→Town02 transfer, semantic-segmentation observations, and a simplified discrete control space, so results should be interpreted as controlled evidence of transfer under town/layout shift rather than a full driving stack. Future work should extend evaluation to richer sensing (e.g., RGB), continuous control, broader towns/weather, and sim-to-real.


\bibliography{main}

\end{document}